\documentclass[conference]{IEEEtran}
\IEEEoverridecommandlockouts

\usepackage[T1]{fontenc}
\usepackage{cite}
\usepackage{amsmath,amssymb,amsfonts,bm}
\usepackage{algorithmic}
\usepackage{graphicx}
\usepackage{textcomp}
\usepackage{xspace}
\usepackage{longtable}
\usepackage{tabularx}
\usepackage{subfigure}
\usepackage{wrapfig}
\usepackage{caption}
\usepackage{graphics}
\usepackage{xcolor}
\usepackage{booktabs}
\def\BibTeX{{\rm B\kern-.05em{\sc i\kern-.025em b}\kern-.08em
    T\kern-.1667em\lower.7ex\hbox{E}\kern-.125emX}}

\usepackage{algorithm}

\newtheorem{theorem}{Theorem}[section]
\newtheorem{corollary}{Corollary}[section]

\newtheorem{proposition}[theorem]{Proposition}
\newtheorem{definition}{Definition}

\usepackage{multibib}

\newcommand{\ouralg}{\texttt{ERL}\xspace}

\newcommand{\opt}{\texttt{OPT}\xspace}
\newcommand{\switch}{\texttt{Switch}\xspace}
\newcommand{\greedy}{\texttt{Robust}\xspace}

\newcommand{\greedyswitch}{\texttt{Greedy}\xspace}
\newcommand{\ml}{\texttt{ML}\xspace}
\newcommand{\rob}{\texttt{ERL-NT}\xspace}    
    
\begin{document}

\title{Robustified Learning for Online Optimization with Memory Costs}

\author{
%\IEEEauthorblockN{1\textsuperscript{st} Given Name Surname}
\IEEEauthorblockN{Pengfei Li}
\IEEEauthorblockA{\textit{UC Riverside}}
\and
\IEEEauthorblockN{Jianyi Yang}
\IEEEauthorblockA{\textit{UC Riverside}}
\and
\IEEEauthorblockN{Shaolei Ren}
\IEEEauthorblockA{\textit{UC Riverside}}
}

\maketitle

\begin{abstract}
Online optimization with memory costs has many real-world applications, where sequential actions are made without knowing the future input. Nonetheless, the memory cost couples the actions over time, adding substantial challenges. Conventionally, this problem has been approached by various expert-designed online algorithms with the goal of achieving bounded worst-case competitive ratios, but the resulting average performance is often unsatisfactory. On the other hand, emerging machine learning (ML) based optimizers can improve the average performance, but suffer from the lack of worst-case performance robustness. In this paper, we propose a novel expert-robustified learning (\ouralg) approach, achieving {both} good average performance and robustness. More concretely, for robustness, \ouralg introduces a novel projection operator that robustifies ML actions by utilizing an expert online algorithm; for average performance, \ouralg trains the ML optimizer based on a recurrent architecture by explicitly considering  downstream expert robustification.  We prove that, for any $\lambda\geq1$, \ouralg can achieve $\lambda$-competitive against the expert algorithm and $\lambda\cdot C$-competitive against the optimal offline algorithm (where $C$ is the expert's competitive ratio). Additionally, we extend our analysis to a novel setting of multi-step memory costs. Finally, our analysis is supported by empirical experiments for an energy scheduling application.
\end{abstract}

\section{Introduction}

Online optimization is a classic sequential decision problem
where the agent chooses irrevocable actions at runtime without knowing the future input. Moreover, in many practical applications, action smoothness over time is highly desired.
For example, for motion planning, a robot cannot move arbitrarily
due to velocity and/or acceleration limitations; for data center capacity provisioning, servers cannot be turned on/off frequently to avoid excessive
wear-and-tear costs and setup delays; and for energy scheduling in smart grids,
quickly adjusting energy production can be very costly \cite{SOCO_OnlineOpt_UntrustedPredictions_Switching_Adam_arXiv_2022,SOCO_DynamicRightSizing_Adam_Infocom_2011_LinWiermanAndrewThereska,SOCO_Memory_Adam_NIPS_2020_NEURIPS2020_ed46558a,Shaolei_L2O_ExpertCalibrated_SOCO_SIGMETRICS_2022}.
Consequently, the long list of real-world applications
have led to the emergence of online optimization with \emph{memory} costs that penalize frequent action changes over time.

Adding a memory cost provides crucial regularization for online action smoothness, but also presents significant algorithmic challenges. More concretely, the memory cost essentially couples the online actions across multiple time steps, making it very challenging, if ever possible, to obtain optimal actions without knowing the future.
Conventionally, this challenge has been approached by expert-designed online algorithms under various settings \cite{SOCO_NonConvex_Adam_Sigmetrics_2020_10.1145/3379484,SOCO_DynamicRightSizing_Adam_Infocom_2011_LinWiermanAndrewThereska,SOCO_Revisiting_Nanjing_NIPS_2021_zhang2021revisiting,SOCO_Memory_Adam_NIPS_2020_NEURIPS2020_ed46558a,SOCO_Prediction_LinearSystem_NaLi_Harvard_NIPS_2019_10.5555/3454287.3455620}. These expert algorithms typically
have worst-case performance robustness in terms of guaranteed competitive ratios
even for adversarial inputs, but
their conservative nature also means that they may not perform very well on average in many typical cases.

More recently, the abundance of historical data  
in practical applications has been fueling machine learning (ML)  approaches to solve optimization problems \cite{L2O_Survey_Amortized_Continuous_Brandon_arXiv_2022_DBLP:journals/corr/abs-2202-00665,Shaolei_L2O_ExpertCalibrated_SOCO_SIGMETRICS_2022,SOCO_OnlineOpt_UntrustedPredictions_Switching_Adam_arXiv_2022}.  
In particular,  optimizers based
on \emph{offline}-trained recurrent neural networks or reinforcement learning
have been emerging for various online optimization problems, including
online resource allocation
\cite{L2O_OnlineResource_PriceCloud_ChuanWu_AAAI_2019_10.1609/aaai.v33i01.33017570},  online knapsack \cite{L2O_NewDog_OldTrick_Google_ICLR_2019}, among others.
These ML-based optimizers exploit the statistical information
about problem inputs and the strong prediction power of neural networks,
empirically achieving unprecedented \emph{average} performance.
But, they also have a significant drawback --- lack of performance robustness. Specifically, unlike expert online algorithms
that have guaranteed robustness,
the competitive ratio of ML-based optimizers can be arbitrarily bad, e.g., when training-testing distributions differ, testing inputs are adversarial, and/or the model capacity is stringently limited \cite{SOCO_OnlineMetric_UntrustedPrediction_ICML_2020_DBLP:conf/icml/AntoniadisCE0S20,SOCO_OnlineOpt_UntrustedPredictions_Switching_Adam_arXiv_2022,Shaolei_L2O_ExpertCalibrated_SOCO_SIGMETRICS_2022}. As a result, the lack of  robustness 
invalidates the existing ML-based optimizers for online optimization in many
real applications, especially those high-stake ones.

To exploit the power of both ML and expert designs,
ML-augmented online algorithms have been recently proposed \cite{OnlineOpt_ML_Augmented_RobustCache_Google_ICML_2021_pmlr-v139-chledowski21a,OnlineOpt_ML_Adivce_Survey_2016_10.1145/2993749.2993766}, including in the context of online optimization with memory costs that we focus on
\cite{SOCO_OnlineMetric_UntrustedPrediction_ICML_2020_DBLP:conf/icml/AntoniadisCE0S20,SOCO_OnlineOpt_UntrustedPredictions_Switching_Adam_arXiv_2022,Shaolei_L2O_ExpertCalibrated_SOCO_SIGMETRICS_2022}. The most common goal of these studies is to  
achieve a finite competitive ratio (i.e., robustness) to bound the worst-case performance for arbitrarily bad
ML outputs
and a low competitive ratio (i.e., consistency) in order to approximately retain good average-case performance enabled by ML models.
Nonetheless, there exist substantial challenges to \emph{simultaneously} 
achieve good robustness and consistency for our problem
setting (see broadly relevant algorithms
\cite{SOCO_OnlineMetric_UntrustedPrediction_ICML_2020_DBLP:conf/icml/AntoniadisCE0S20,SOCO_OnlineOpt_UntrustedPredictions_Switching_Adam_arXiv_2022,Shaolei_L2O_ExpertCalibrated_SOCO_SIGMETRICS_2022}), let alone that 
 a good consistency may not always translate into a good average performance. 
Moreover,
the existing ML-augmented algorithms often view the ML model as an exogenous blackbox that is pre-trained as a standalone model without being aware of the downstream expert  algorithm. This essentially creates a \emph{mismatch} between training and testing --- the ML model is trained alone but tested together with a downstream algorithmic procedure
--- which can unnecessarily hurt the resulting average performance.

In this paper, we focus on online optimization with memory costs
and propose a novel expert-robustified learning (\ouralg) approach, achieving
{both} good average performance and guaranteed robustness. The key idea of \ouralg is to \emph{let the expert and ML do what they are best at respectively}:
for guaranteed robustness, \ouralg utilizes an expert online algorithm to
robustify the ML actions by projecting them into a carefully designed robust action space; for good average performance, \ouralg trains the ML model by explicitly considering the downstream expert robustification process, thus avoiding the mismatch between training and testing.
We prove that, for any trust hyperparameter $\lambda\geq1$ governing how much flexibility we allow for ML actions, \ouralg can achieve $\lambda$-competitive against the expert online algorithm and hence $\lambda\cdot C$-competitive against the optimal offline algorithm (where $C$ is the expert's competitive ratio).
The added  robustification step is an implicit layer, making it non-trivial to perform backpropagation. Thus, we also derive gradients of the robustification step with respect to their inputs for efficient end-to-end training, thus improving the average-case performance. We subsequently extend our analysis to a novel setting, where the memory cost spans multiple steps. Finally, we run experiments to empirically validate \ouralg for an energy scheduling application, demonstrating that it can offer the best
average cost and competitive ratio tradeoff.

\section{Related Works}

Online optimization with (single-step) memory costs has been extensively approached under various settings by  expert algorithms, such as online gradient descent (OGD) \cite{OGD_zinkevich2003online}, online balanced descent (OBD) \cite{SOCO_OBD_Niangjun_Adam_COLT_2018_DBLP:conf/colt/ChenGW18}, and regularized OBD (R-OBD) \cite{SOCO_OBD_R-OBD_Goel_Adam_NIPS_2019_NEURIPS2019_9f36407e}.
Additionally, expert algorithms with the knowledge of future inputs
include receding horizon control (RHC) \cite{SOCO_Prediction_Error_Meta_ZhenhuaLiu_SIGMETRICS_2019_10.1145/3322205.3311087}
committed horizon control (CHC) \cite{SOCO_Prediction_Error_Niangjun_Sigmetrics_2016_10.1145/2964791.2901464},
receding horizon gradient descent (RHGD) \cite{Receding_Horizon_GD_li2020online,SOCO_Prediction_LinearSystem_NaLi_Harvard_NIPS_2019_10.5555/3454287.3455620}.
These algorithms are judiciously
designed to have bounded competitive ratios and/or regrets, but they may not perform well on average.

ML-augmented algorithm designs have also been emerging in the context of online optimization with memory costs \cite{SOCO_OnlineMetric_UntrustedPrediction_ICML_2020_DBLP:conf/icml/AntoniadisCE0S20,SOCO_OnlineOpt_UntrustedPredictions_Switching_Adam_arXiv_2022}. 
Nonetheless, these algorithms simply take the actions produced
by an exogenous ML-based optimizer as additional inputs; they still focus on
on manual designs, which cannot achieve good worst-case and average performance
simultaneously. For example, in order to retain the good average performance
of ML actions by setting a hyperparameter $\delta\to0$, the competitive ratio when ML actions are arbitrarily bad
is as high as $\frac{12+o(1)}{\delta}\left(\frac{2}{\alpha+\delta(1+\alpha)}\right)^{{2}/{(\delta\alpha)}}$ for $\alpha$-polyhedral cost functions \cite{SOCO_OnlineOpt_UntrustedPredictions_Switching_Adam_arXiv_2022}. 
The study \cite{Shaolei_L2O_ExpertCalibrated_SOCO_SIGMETRICS_2022} considers a squared single-step switching cost and trains an ML model to regularize online actions, but its worst-case competitive ratio is unbounded. In orthogonal contexts,
by assuming a given downstream algorithm,
\cite{L2O_LearningMLAugmentedAlgorithm_Harvard_ICML_2021_pmlr-v139-du21d}
re-trains an ML model
for the count-min sketch problem.
Therefore, the novel expert robustification (for tunable and bounded performance robustness), end-to-end training (for good average performance), and new problem settings altogether separate our work far apart from the literature.

Learning to optimize (L2O) based on offline-trained recurrent neural networks
or reinforcement learning  \cite{L2O_LearningToOptimize_Berkeley_ICLR_2017} has been recently applied for online optimization, including online resource allocation and online bipartite matching \cite{L2O_OnlineResource_PriceCloud_ChuanWu_AAAI_2019_10.1609/aaai.v33i01.33017570,L2O_NewDog_OldTrick_Google_ICLR_2019}. Nonetheless,
even with the help of adversarial training \cite{L2O_AdversarialOnlineResource_ChuanWu_HKU_TOMPECS_2021_10.1145/3494526},
a crucial drawback of the existing ML-based optimizers is the lack of guaranteed performance robustness, making them inapplicable for high-stake applications. Naive techniques that choose whichever is better between
L2O and a conventional solver \cite{L2O_Safeguard_SimpleConvex_ZhangyangWang_arXiv_2020_https://doi.org/10.48550/arxiv.2003.01880}  do not apply to online optimization due to unknown future inputs and irrevocable actions.

\ouralg is relevant to the recent  decision-focused learning framework 
\cite{L2O_PredictOptimize_MDP_Harvard_NIPS_2021_wang2021learning}.
But, \ouralg goes beyond
simply training the ML model by proposing a novel
expert robustification framework.
Moreover, \ouralg directly uses the robustified actions to determine the training loss, whereas the existing decision-focused learning requires groundtruth labels in the training loss.

Finally, \ouralg  intersects with  
conservative exploration in bandits and reinforcement learning \cite{Conservative_RL_Bandits_LiweiWang_SimonDu_ICLR_2022_yang2022a}. Conservative exploration focuses on unknown
reward functions (and transition models if applicable) and
uses an existing policy to guide the exploration process for robustness. But, its design is dramatically different in the sense that it does not need to account for future input uncertainties when making an action for each step (or choosing a policy for each episode in case of episodic reinforcement learning), i.e., only the cumulative rewards matter. By contrast, \ouralg must hold a reservation cost to ensure that it always has a feasible solution given any future inputs, achieving a guaranteed deterministic worst-case competitive ratio (rather than probabilistic guarantees). This key point can also be highlighted by noting that, even assuming perfect reward functions (and transition models), the robustification rule used by the existing conservative bandits/reinforcement learning \cite{Conservative_RL_Bandits_LiweiWang_SimonDu_ICLR_2022_yang2022a}
cannot apply to our problem to achieve a guaranteed competitive ratio. 
Other related problems include constrained policy optimization
and safe reinforcement learning \cite{Conservative_ProjectBasedConstrainedPolicyOptimizatino_ICLR_2020_Yang2020Projection-Based,Conservative_RL_OfflineDistributional_JasonMa_NIPS_2021_ma2021conservative}. These studies focus on constraining the \emph{average} safety costs and/or avoiding certain dangerous states (possibly with a high probability). By contrast, \ouralg has a different goal and guarantees a bounded competitive ratio in \emph{any} case by introducing a novel expert robustification step.

\section{Formulation for Single-Step Memory Cost}

To facilitate readers' understanding,
we begin with a single-step memory cost (a.k.a. switching cost \cite{SOCO_DynamicRightSizing_Adam_Infocom_2011_LinWiermanAndrewThereska,SOCO_NonConvex_Adam_Sigmetrics_2020_10.1145/3379484,SOCO_Revisiting_Nanjing_NIPS_2021_zhang2021revisiting}). Consider a sequence of $T$ time steps as a problem instance.
At each step $t=1,\cdots,T$, the agent receives a context vector/parameter $y_t\in\mathcal{Y}\in\mathbb{R}^m$ for the hitting cost, makes an \emph{irrevocable} action $x_t\in\mathcal{X}\subseteq\mathbb{R}^d$, and then incurs a hitting cost of $f(x_t,y_t)\geq0$. To encourage smoothed actions over time, the agent also incurs a memory cost $d(x_t,x_{t-1})\geq0$ defined in terms of the distance between two adjacent actions in a metric space.
Concretely, we consider $d(x_t,x_{t-1})=\|x_t-x_{t-1}\|$, where $\|\cdot\|$ denotes $l_p$ norm with $p\geq1$. Thus, the goal of the agent is to minimize the sum of the hitting costs and the memory costs over a sequence of $T$ steps as follows:
\begin{equation}\label{eqn:naive_online}
	\min_{x_1,\cdots x_T} \sum_{t=1}^Tf(x_t,y_t)+d(x_t,x_{t-1}),
\end{equation}
where the initial action $x_0$ is provided as an additional input. While we can alternatively impose a constraint on the total memory cost,
our formulation of adding the memory cost as a smoothness regularizer for online actions is consistent with the existing literature  \cite{SOCO_DynamicRightSizing_Adam_Infocom_2011_LinWiermanAndrewThereska,SOCO_Revisiting_Nanjing_NIPS_2021_zhang2021revisiting,SOCO_OnlineOpt_UntrustedPredictions_Switching_Adam_arXiv_2022,Shaolei_L2O_ExpertCalibrated_SOCO_SIGMETRICS_2022}.

The key challenge for optimally solving Eqn.~\eqref{eqn:naive_online} comes from
the memory cost that couples online actions over time,
but 
$y_t$
is not revealed to the agent until
the beginning of each step $t=1,\cdots,T$.
Given any online algorithm $\pi$,  
we denote its total
cost for a problem instance with input context $\bm{s}=(x_0,\bm{y})\in\mathcal{S}=\mathcal{X} \times \mathcal{Y}^T$
as $\mathrm{cost}(\pi,\bm{s})=\sum_{t=1}^Tf(x_t^\pi,y_t)+d(x_t^\pi,x_{t-1}^\pi)$, where $x_t^\pi$, $t=1,\cdots, T$,
 are the actions produced by the algorithm $\pi$.
 While $\bm{s}$ follows a general distribution that can be well addressed
by ML-based optimizers, it can still contain adversarial
cases. 
For simplicity, we will 
omit the context parameters,
and
denote $\mathrm{cost}(x_{i:j}^{\pi})=\sum_{t=i}^jf(x_t^\pi,y_t)+d(x_t^\pi,x_{t-1}^\pi)$, where
$x_{i:j}^{\pi}=(x_i^{\pi},\cdots, x_j^{\pi})$ are the actions for $t=i,\cdots,j$
under the algorithm $\pi$.

\begin{definition}[\bf $\alpha$-polyhedral]
\label{def:polyhedral}
Given a context parameter $y\in\mathcal{Y}$, the hitting cost function $f(x,y): \mathcal{X}\mapsto \mathbb{R}^+$ is called $\alpha$-polyhedral for $\alpha>0$
if it has a unique minimizer $x^*\in\mathcal{X}$ and satisfies 
$f(x,y) - f(x^*,y) \geq \alpha\cdot d(x,x^*)$
for any $x\in\mathcal{X}$.
\end{definition}

\begin{definition}[\bf Competitive ratio]
\label{def:cr}
For $\lambda\geq1$, an online algorithm $ALG$ is called $\lambda$-competitive against the algorithm $\pi$ subject to an additive
factor $B\geq0$ 
if its total cost satisfies
$\mathrm{cost}(ALG,\bm{s})\leq \lambda\cdot\mathrm{cost}(\pi,\bm{s}) + B,
$ for any input $\bm{s}=(x_0,\bm{y})$.
\end{definition}

The $\alpha$-polyhedral definition is commonly considered
in the literature \cite{SOCO_OnlineOpt_UntrustedPredictions_Switching_Adam_arXiv_2022,SOCO_Revisiting_Nanjing_NIPS_2021_zhang2021revisiting} to derive competitive ratios against the optimal offline algorithm. The deterministic competitive ratio
in Definition~\ref{def:cr} is general, and
the additive factor $B$ is independent of
the problem input $\bm{s}=(x_0,\bm{y})$.
By setting %the additive factor
$B=0$,
it becomes 
the strict competitive ratio \cite{SOCO_OnlineOpt_UntrustedPredictions_Switching_Adam_arXiv_2022,OnlineConvexOpt_Book_2016_hazan2016introduction,OnlineMatching_Booklet_OnlineMatchingAdAllocation_Mehta_2013_41870}. Further, with $\lambda=1$, the additive factor $B$ 
in Definition~\ref{def:cr}
 captures the \emph{regret} incurred by $ALG$ 
with respect to the algorithm $\pi$.
When $\pi$ is not specified, 
the competitive ratio
is against the optimal offline
algorithm  \opt by default.

\section{\ouralg: Expert-Robustified Learning}\label{sec:algorithm}

In this section, we consider a single-step memory cost and show the design of \ouralg.

\subsection{A Primer on Pure ML-based Optimizers}
To solve online optimization with memory costs,
a natural idea is to exploit the power of ML to discover
the mapping from the available online information to  
actions. More concretely, we can pre-train an ML model offline
based on a recurrent neural
network (RNN) or equivalently using reinforcement learning. We denote
the ML action at time $t$ as $\tilde{x}_t=h_w(\tilde{x}_{t-1},y_t)$, where $w$ is the ML model parameter. The recurrent nature comes from sequential online optimization with memory costs: given the previous action $\tilde{x}_{t-1}$ and the current  input $y_t$, we recurrently output an online action $\tilde{x}_t$.
With a set of training problem instances, the ML model parameter $w$ can be learnt by minimizing a loss function, which can be %either 
the sum of costs in Eqn.~\eqref{eqn:naive_online}
 \cite{L2O_Survey_Amortized_Continuous_Brandon_arXiv_2022_DBLP:journals/corr/abs-2202-00665}.

\textbf{Drawbacks:} 
It is well-known that such ML-based optimizers have significant drawbacks --- lack of robustness. 
Specifically, the competitive ratio can be
 arbitrarily bad for a variety of reasons, such as
 distributional shifts, hard problem instances or even adversarial inputs, and/or finite ML capacity \cite{SOCO_OnlineMetric_UntrustedPrediction_ICML_2020_DBLP:conf/icml/AntoniadisCE0S20,SOCO_OnlineOpt_UntrustedPredictions_Switching_Adam_arXiv_2022,L2O_LearningMLAugmented_Regression_CR_GeRong_NIPS_2021_anand2021a}. 
While distributionally robust learning can partially mitigate the lack
of robustness in an average sense \cite{DRO_LabelShiftDRO_ICLR_2021_zhang2021coping,DRO_MMD_staib2019distributionally}, it still cannot guarantee that the ML model has a bounded
competitive ratio for \emph{any} problem instance.

\subsection{Expert Robustification}\label{sec:robustification}

There have been several expert algorithms to solve online optimization with memory costs under different settings \cite{SOCO_Revisiting_Nanjing_NIPS_2021_zhang2021revisiting,SOCO_Prediction_Error_RHIG_NaLi_Harvard_NIPS_2020_NEURIPS2020_a6e4f250}.
While these algorithms may not perform well on average due to their conservative nature,
they offer worst-case performance robustness for any  input.  
Thus, this motivates us to leverage an expert algorithm $\pi$ to robustify ML actions. For each $t$, we denote the pre-robustification ML action as $\tilde{x}_t$, expert action as $x_t^{\pi}$, and post-robustification action as $x_t$.

A naive idea is to add a proper regularizer during the training process
that imposes penalty
when the total cost exceeds $\lambda$ times of the expert's cost.
But, this will not work, because the ML actions can still violate the robustness
requirement when bad problem instances arrive
during online inference. Alternatively, one may want to constrain the robustified actions such that for any $t=1,\cdots, T$, the cumulative cost up to time $t$ satisfies $\text{cost}(x_{1:t}) \leq \lambda\text{cost}(x_{1:t}^\pi) + B$, where
$\text{cost}(x_{1:t})$ and $\text{cost}(x_{1:t}^\pi)$ are the cumulative costs of \ouralg and the expert (assuming that the expert would run its algorithm alone), respectively. But, this can easily result in an empty set of feasible actions for \ouralg, because the actions are coupled over time by memory costs. To see this point, let us consider that $\text{cost}(x_{1:t}) = \lambda\text{cost}(x_{1:t}^\pi) + B$ but $x_t\not=x_t^{\pi}$ at time $t$. Then, at time $t+1$, the expert can have such a low total cost of $f(x_{t+1}^{\pi},y_{t+1})+d(x_{t+1}^{\pi},x_{t}^{\pi})$ that even setting $x_{t+1}=x_{t+1}^{\pi}$ (i.e., following the expert) 
would violate
the constraint $\text{cost}(x_{1:t+1}) \leq \lambda\text{cost}(x_{1:t+1}^\pi) + B$
due to the large memory cost $d(x_{t+1}^{\pi},x_t)$. Consequently,
no actions can guarantee robustness in this case.

We now present our novel robustification framework, called \ouralg. To achieve robustness, the crux of \ouralg is to hedge against
the risk of deviating from the expert action to account for future
uncertainty. Specifically, at each step $t$, we project
the ML action $\tilde{x}_t$ into a robust action space specified by the expert $\pi$ by solving:
\begin{equation}
    \begin{split}
    \begin{gathered}
    \label{eqn:proj_naive}
    x_t  = \arg\min_{x\in\mathcal{X}} \frac{1}{2} \| x - \tilde{x}_t \| ^2\\
    s.t. \quad  \text{cost}(x_{1:t-1}) + f(x,y_t) + d(x, x_{t-1})  +  d(x,x_t^\pi) \\
    \leq \lambda\text{cost}(x_{1:t}^\pi) + B
    \end{gathered}
    \end{split}
\end{equation}
where $\lambda\geq1$ and $B\geq 0$ 
are hyperparameters indicating
the level of robustness requirement. We denote this projection step as $x_t = \text{proj}(\tilde{x}_t, x_t^\pi, \text{cost}(x_{1:t-1}),
\text{cost}(x_{1:t}^\pi))$. Importantly, the key
is to add a reservation cost $d(x_t,x_t^{\pi})$ when constraining the post-robustification cumulative cost at time $t$ in Eqn.~\eqref{eqn:proj_naive}. By doing so, we ensure that if the constraint is satisfied
at time $t$, then it will be also satisfied at time $t+1$ regardless of the input --- following the expert by choosing $x_{t+1}=x_{t+1}^{\pi}$ is always a 
\emph{feasible} solution.

\begin{figure}[!t]
    \includegraphics[trim=0 0cm 0 0, clip, width=0.47\textwidth]{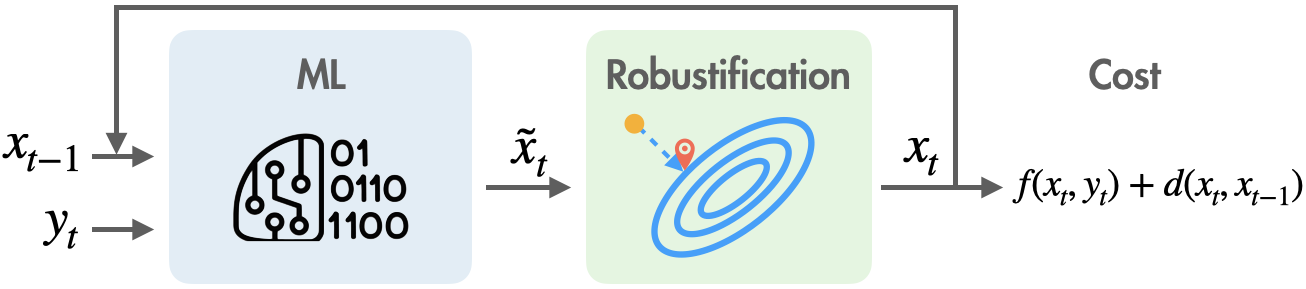}
    %\vspace{-0.6cm}
    \caption{\ouralg. Given each online input,
     we first run forward inference to obtain
     the ML action, and then project it into an expert-robusitfied action space
     as the actual action.}
    \label{fig:model_rnn}
\end{figure}
The  \ouralg inference process is shown in Fig.~\ref{fig:model_rnn}
and described in Algorithm~\ref{alg:EROP}.
At each step $t$, we run the ML model to produce an action $\tilde{x}_t$, get the expert's action $x_t^{\pi}$, and then project 
$\tilde{x}_t$ into a robustified 
action space by solving Eqn.~\eqref{eqn:proj_naive}.
Note that the expert online algorithm takes context $y_t$ as its input
and outputs its action 
$x_t^{\pi}$ independently  following its own trajectory without being affected by the ML action. Next, we formally provide the robustness analysis for \ouralg.

\begin{algorithm}[t!]
\caption{Expert-Robustified Learning
(\ouralg)}
\begin{algorithmic}[1]\label{alg:EROP}
\REQUIRE $\lambda\geq1$, $B \geq 0$,
initial $x_0$, trained ML model (Section~\ref{sec:training}), and expert online algorithm $\pi$
\STATE for $t=1,\cdots, T$
\STATE \quad Receive the context $y_t$
\STATE \quad Expert chooses $x_t^{\pi}$ and
ML chooses $\tilde{x}_t \leftarrow h(x_{t-1}, y_t)$ 
\STATE \quad $x_t \leftarrow \text{proj}(\tilde{x}_t, x_t^\pi, \text{cost}(x_{1:t-1}), \text{cost}(x_{1:t}^\pi))$ based on Eqn.~\eqref{eqn:proj_naive} \quad  \texttt{//Robustification}
\end{algorithmic}
\end{algorithm}

\begin{theorem}\label{theorem:cr_online} Let $\pi$ be any expert online algorithm for the 
problem in Eqn.~\eqref{eqn:naive_online}.
For any $\lambda\geq1$ and $B\geq0$,
\ouralg is $\lambda$-competitive against $\pi$ subject
to an additive factor of $B$,
i.e.,
$\mathrm{cost}(\ouralg,\bm{s})\leq \lambda\cdot\mathrm{cost}(\pi,\bm{s}) + B
$ for any input $\bm{s}=(x_0,\bm{y})$.
\end{theorem}

Theorem~\ref{theorem:cr_online} is proved
in Appendix~\ref{sec:proof_E2}
and demonstrates the power of \ouralg 
by showing that
it can achieve any competitive ratio of $\lambda\geq1$ with respect to any expert algorithm $\pi$ for $B\geq0$.
Here, given any $B\geq0$, the hyperparameter $\lambda\geq1$ 
can be viewed as the \emph{trust} parameter: the higher $\lambda$, the more we trust the ML action $\tilde{x}_t$, thus potentially achieving
a lower average cost at the expense of a higher competitive ratio.
The additive factor $B\geq0$ represents a slackness,
and $B=0$ reduces to the strict competitive ratio definition.  
If we set $\lambda=1$
and $B$ sublinear in $T$, \ouralg is guaranteed %no-regret learning 
to be asymptotically no worse than the expert $\pi$ even in the worst case as $T\to\infty$.

\textbf{Competitive ratio
of \ouralg against \opt.} 
One may also desire a bounded competitive ratio against the optimal offline algorithm \opt. To this end, we consider a state-of-the-art expert
online algorithm, called \greedy, which minimizes the hitting cost
at each step without considering the memory cost. This simple online algorithm
surprisingly achieves a good competitive ratio of $\max\left(\frac{2}{\alpha},1\right)$ against \opt for $\alpha$-polyhedral hitting cost functions \cite{SOCO_Revisiting_Nanjing_NIPS_2021_zhang2021revisiting}. 
By applying
Theorem~\ref{theorem:cr_online}, we have the following corollary.

\begin{corollary}\label{corollary:cr_opt} Consider \greedy 
as the algorithm $\pi$ that chooses
$x_t=\arg\min_{x\in\mathcal{X}}f(x,y_t)$ for any $t$. Assume
that the hitting cost functions
$f(x,y_t)$
are $\alpha$-polyhedral.
 For any $\lambda\geq1$ and $B\geq0$,
\ouralg is
$\lambda\cdot \max\left(\frac{2}{\alpha},1\right)$-competitive
against %the optimal offline algorithm 
\opt subject
to an additive factor of $B$,
i.e.,
$\mathrm{cost}(\ouralg,\bm{s})\leq \lambda \cdot\max\left(\frac{2}{\alpha},1\right)\cdot\mathrm{cost}(\opt,\bm{s}) + B
$ for any input $\bm{s}=(x_0,\bm{y})$.
\end{corollary}

\subsection{End-to-end Training}\label{sec:training}

In conventional ML-augmented algorithms \cite{SOCO_ML_ChasingConvexBodiesFunction_Adam_COLT_2022,SOCO_OnlineMetric_UntrustedPrediction_ICML_2020_DBLP:conf/icml/AntoniadisCE0S20},
the ML model is trained to produce good actions on its own,
without being aware of the downstream modification (i.e., expert robustification in \ouralg).
While the designed algorithm may sometimes retain good ML actions (i.e., termed as {consistency} \cite{SOCO_OnlineMetric_UntrustedPrediction_ICML_2020_DBLP:conf/icml/AntoniadisCE0S20}),
this still creates a mismatch between training and testing processes --- 
the training process yields good \emph{pre}-robustification ML actions,
but it is \emph{post}-robustification
actions that are actually being used for testing
\cite{Shaolei_L2O_ExpertCalibrated_SOCO_SIGMETRICS_2022,L2O_Survey_Amortized_Continuous_Brandon_arXiv_2022_DBLP:journals/corr/abs-2202-00665,L2O_PredictOptimize_RiskCalibrationBound_ICLR_2021_liu2021risk,L2O_PredictOptimize_MDP_Harvard_NIPS_2021_wang2021learning}. 
Thus, to improve the \emph{average} performance
of \ouralg, we need to explicitly consider the projection step for ML model training. 

End-to-end training is highly non-trivial,
since the projection step itself is an optimization problem in Eqn.~\eqref{eqn:proj_naive} and hence an \emph{implicit} layer. Additionally, unlike typical differentiable optimizers \cite{L2O_DifferentiableConvexOptimization_Brandon_NEURIPS2019_9ce3c52f,L2O_DifferentiableMPC_NIPS_2018_NEURIPS2018_ba6d843e,L2O_PredictOptimize_MDP_Harvard_NIPS_2021_wang2021learning}, we
need to derive gradients to perform backpropagation through time due to the recurrent nature of our online optimization problem. Let $w$ be the weight for each base ML model 
$\tilde{x}_t=h_w(x_{t-1},y_t)$
in the RNN as illustrated in Fig.~\ref{fig:model_rnn}.
 We need to derive the gradient of $\text{cost}(x_{1:T})$ with respect to $w$ as follows: 
 %which is expressed as
\begin{equation}
   \nabla_w \text{cost}(x_{1:T}) = \sum_{t = 1}^T \nabla_w \big( f(x_t, y_t) + d(x_t, x_{t-1}) \big), 
\end{equation}
where $x_t=\text{proj}(\tilde{x}_t, x_t^\pi, \text{cost}(x_{1:t-1}), \text{cost}(x_{1:t}^\pi))$ is the post-robustification action
at step $t=1,\cdots,T$.
Thus, the total gradient can be calculated by summing up all the gradients over $T$ steps. 
By applying the chain rule for step $t$, 
we have
$\nabla_w \big( f(x_t, y_t) + d(x_t, x_{t-1}) \big) 
    =  \nabla_{x_t} \big( f(x_t, y_t) + d(x_t, x_{t-1}) \big)\nabla_w {x_t}  + \nabla_{x_{t-1}} d(x_t, x_{t-1}) \cdot \nabla_w {x_{t-1}}$,
where $\nabla_w {x_t} = \left( \nabla_{\tilde{x}_t} x_t \nabla_w \tilde{x}_t + \nabla_{\text{cost}(x_{1:t-1})} x_t \nabla_w \text{cost}(x_{1:t-1}) \right)$.
 The gradients $\nabla_{x_t} \big( f(x_t, y_t) + d(x_t, x_{t-1}) \big)$ 
 and $\nabla_{x_{t-1}} d(x_t, x_{t-1})$
 can be obtained given explicit forms of 
$f$ and $d$, $\nabla_w \tilde{x}_t$ can be calculated easily through the backpropagation within the ML model
(e.g., a neural network), and  $\nabla_w \text{cost}(x_{1:t-1})$ is calculated recursively back to $t =1$. %For $t=1$, $\nabla_w \text{cost}(x_{1:t-1}) = 0$.
Thus, the key is to derive the gradients of the projection operator
$x_t=\text{proj}(\tilde{x}_t, x_t^\pi, \text{cost}(x_{1:t-1}), \text{cost}(x_{1:t}^\pi))$ with respect to the ML action $\tilde{x}_t$ and $\text{cost}(x_{1:t-1})$.
We provide the result
based on KKT conditions \cite{BoydVandenberghe} in the following proposition. 

\begin{proposition}[Gradient by KKT conditions]\label{thm:back-propogation}
 Assume that $x_t$ and $\mu$ are the primal and dual solutions to Eqn.~\eqref{eqn:proj_naive}, respectively. Let $\Delta_{11} = I + \mu \Big( \nabla_{x_t,x_t} \big( f(x_t, y_t) + d(x_t, x_{t-1}) \big) + \nabla_{x_t,x_t} d(x_t, x_t^\pi) \Big)$, $\Delta_{12} = \nabla_{x_t} \big( f(x_t, y_t) + d(x_t, x_{t-1}) \big) + \nabla_{x_t} d(x_t, x_t^\pi)$, $\Delta_{21} = \mu \Big( \nabla_{x_t} \big( f(x_t, y_t) + d(x_t, x_{t-1}) \big) + \nabla_{x_t} d(x_t, x_t^\pi) \Big)^\top$, $\Delta_{22} = f(x_t, y_t) + d(x_t, x_{t-1}) + d(x_t,x_t^\pi) + \text{cost}(x_{1:t-1}) -  \left[ \lambda\text{cost}(x_{1:t}^\pi ) + B \right]$. 
 The gradients of the projection operation
 $x_t=\text{proj}(\tilde{x}_t, x_t^\pi, \text{cost}(x_{1:t-1}), \text{cost}(x_{1:t}^\pi))$ with respect to $\tilde{x}_t$ and $\text{cost}(x_{1:t-1})$ are
$$\nabla_{\tilde{x}_t} x_t =  \Delta_{11}^{-1} [I + \Delta_{12} Sc(\Delta,\Delta_{11})^{-1}\Delta_{21} \Delta_{11}^{-1} ],$$
$$\nabla_{\text{cost}(x_{1:t-1})} x_t = \Delta_{11}^{-1} \Delta_{12} Sc(\Delta, \Delta_{11})^{-1} \mu,$$
where $Sc(\Delta, \Delta_{11})=\Delta_{22}-\Delta_{21}\Delta^{-1}_{11}\Delta_{12}$ is the Schur-complement of $\Delta_{11}$ in the blocked matrix $\Delta = \big[[\Delta_{11}, \Delta_{12}],[\Delta_{21}, \Delta_{22}] \big]$.
\end{proposition}

We remark that 
if the Schur-complement $Sc(\Delta, \Delta_{11})$ is not full-rank
(e.g., ML action $\tilde{x}_t$ lies in the boundary of
the action space in Eqn.~\eqref{eqn:proj_naive})
 or
 the hitting cost function
$f$ or memory cost $d$ is {not 
differentiable for certain $x_t$}, 
we can still approximate the gradients based on Proposition~\ref{thm:back-propogation} for backpropagation. Concretely, 
the pseudo-inverse of $Sc(\Delta, \Delta_{11})$ can be used if 
$Sc(\Delta, \Delta_{11})$ is not full-rank; if $f$ or $d$ is not differentiable at $x_t$,
we can use its its subgradient as a substitute. This is also a common technique to handle non-differentiable points when training ML models, especially neural networks \cite{DNN_Book_Goodfellow-et-al-2016}.
For example, we often use $0$ 
as a subgradient for $ReLu(x)$
at $x=0$. Importantly, 
Proposition~\ref{thm:back-propogation}
provides a practically convenient way to 
perform backpropagation.

\textbf{Training.} As in typical  ML-based
approaches for online optimization \cite{L2O_Survey_Amortized_Continuous_Brandon_arXiv_2022_DBLP:journals/corr/abs-2202-00665,L2O_NewDog_OldTrick_Google_ICLR_2019,L2O_OnlineBipartiteMatching_Toronto_ArXiv_2021_DBLP:journals/corr/abs-2109-10380,L2O_AdversarialOnlineResource_ChuanWu_HKU_TOMPECS_2021_10.1145/3494526}, 
we train the ML model based on pre-collected historical problem instances by
using the gradients in Proposition~\ref{thm:back-propogation} and
explicitly considering the projection process. Additionally, we can 
also update the ML model online by collecting batches of new problem instances 
during online inference. 
The training process can be supervised by using the total cost $\sum_{i}\mathrm{cost}_i(x_{1:T})$ as the loss where $i$ is the index for training problem instances.

\section{Extension to Multi-Step Memory Cost}\label{sec:multi_step_memory}

Motivated by smoothness in higher-order dynamics, we now turn to a more general case where the memory cost can span multiple
steps: 
$d(x_t, x_{t-q:t-1})=\|x_t - \sum_{i=1}^q C_i x_{t-i}\|$,
where  $q\geq1$ is the memory length and $C_i \in \mathbb{R}^{d \times d}$ is problem-specific. For example, let us consider a robot motion planning problem where
$x_t$ represents the position at time $t$ and acceleration smoothness is highly desired. In this case,
the memory cost can be written as
$d(x_t,x_{t-2:t-1})=\|\left(x_t - x_{t-1}
\right) - \left(x_{t-1}- x_{t-2}\right)\|=\|x_t -2 x_{t-1} + x_{t-2}\|$, for which we can set
$C_1 = -2\cdot I$, $C_2 = I$ and $q=2$
where $I$ is the identity matrix in $\mathbb{R}^{d \times d}$. Note that the expert algorithm
in \cite{SOCO_Memory_Adam_NIPS_2020_NEURIPS2020_ed46558a} uses the same form of multi-step memory structure, but considers a \emph{squared} memory cost
along with other strong assumptions (e.g., strongly convex hitting costs)
that require entirely different techniques \cite{SOCO_Revisiting_Nanjing_NIPS_2021_zhang2021revisiting}. To our knowledge, our work
is the first to consider multi-step memory
costs in metric space.

\textbf{Expert robustification.}
Given multi-step memory costs, the input to our ML model includes
$y_t$ and $x_{t-q:t-1}$ and outputs $\tilde{x}_t$, which is then robustified by solving the following:
\begin{equation}\label{eqn:proj_ho}
    \begin{split} 
    \begin{gathered}
    x_t =  \arg\min_{x\in\mathcal{X}} \frac{1}{2}\|x - \tilde{x}_t \|^2 \\
     s.t. \;\;\;\;\;
     \mathrm{cost}({x}_{1:t-1}) + f(x, y_t) + d(x,   x_{t-q:t-1})  \\
     + G(x, {x}_{t-q:t-1}, {x}_{t-q:t}^{\pi})   \leq \lambda\mathrm{cost}({x}_{1:t}^{\pi}) + B,
     \end{gathered}
    \end{split}
\end{equation}
where the reservation cost $G(x, {x}_{t-q:t-1}, {x}_{t-q:t}^{\pi})$ 
is given by 
\begin{equation}\label{eqn:reservation_highorder} 
\begin{split}
&G\left(x, {x}_{t-q:t-1}, {x}_{t-q:t}^{\pi}\right)\\ 
=& \sum_{k = 1}^{\min(q,T-t)}{\left\|C_{k} x +  \sum_{i=1}^{q-k} C_{k + i}x_{t-i}-  \sum_{i=0}^{q-k} C_{k + i}x_{t-i}^{\pi} \right\| }.
\end{split}
\end{equation}
The key insight for Eqn.~\eqref{eqn:reservation_highorder} is
that we need to account for the potentially higher memory costs
incurred by \ouralg compared to the expert algorithm $\pi$ 
over up to future $q$ steps. By holding the reservation cost for
the cumulative cost at each step, we can ensure that \ouralg can always roll back
to the expert's actions in the future without violating the robustness requirement. The \ouralg inference process still follows Algorithm~\ref{alg:EROP}, except for that the projection step for expert robustification in Line~5 is based
on Eqn.~\eqref{eqn:proj_ho}. 

\textbf{Competitive ratio of \greedy.}
\greedy has a bounded competitive ratio
in the single-step memory setting
\cite{SOCO_Revisiting_Nanjing_NIPS_2021_zhang2021revisiting},
but it is unclear in the multi-step setting. 
Here, we prove that \greedy is also competitive in the multi-step memory case.
The proof is in Appendix~\ref{sec:greedy_porrf}.
\begin{theorem}\label{thm:greedy_multi}
Assume that $f(\cdot, y_t) : \mathcal{X} \mapsto \mathbb{R}$ is $\alpha$-polyhedral
and that the memory cost is given by $d(x_t,x_{t-q:t-1})=\|x_t - \sum_{i=1}^q C_i x_{t-i}\|$ for $t=1,\cdots, T$, where $C_i \in \mathbb{R}^{d \times d}$ and $\sum_{i=1}^q \|C_i\| = \beta$ with $\|C_i\|$ being the matrix norm induced by the $l_p$ vector norm. 
The \greedy algorithm that chooses
$x_t=\arg\min_{x\in\mathcal{X}}f(x,y_t)$ for any $t=1,\cdots,T$
is strictly $\max\left(\frac{\beta + 1}{\alpha}, 1\right)$-competitive against \opt, i.e., $\mathrm{cost}(\greedy,\bm{s})\leq \max\left(\frac{\beta + 1}{\alpha}, 1\right)\cdot\mathrm{cost}(\opt,\bm{s}) 
$ for any input $\bm{s}=(x_0,\bm{y})$.
\end{theorem}

\textbf{Competitive ratio of \ouralg.}
In the multi-step memory case, Theorem~\ref{theorem:cr_online} still holds. That is, for any $\lambda\geq1$ and $B\geq0$, \ouralg is still $\lambda$-competitive against any expert online algorithm $\pi$ subject to an additive factor $B$.
Also, by combining this result with Theorem~\ref{thm:greedy_multi}, we obtain the following corollary (proof in Appendix~\ref{sec:proof_E2}).

\begin{corollary}\label{corollary:cr_opt_multi} Let the expert $\pi$ be \greedy that chooses
$x_t=\arg\min_{x\in\mathcal{X}}f(x,y_t)$ for any $t=1,\cdots,T$. Under the same assumptions as in Theorem~\ref{thm:greedy_multi},
 for any $\lambda\geq1$ and $B\geq0$,
\ouralg is
$\lambda\max\left(\frac{\beta+1}{\alpha},1\right)$-competitive
against 
\opt subject
to an additive factor of $B$ where
$\beta=\sum_{i=1}^q \|C_i\|$,
i.e.,
$\mathrm{cost}(\ouralg,\bm{s})\leq \lambda\max\left(\frac{\beta+1}{\alpha},1\right)\cdot\mathrm{cost}(\opt,\bm{s}) + B
$ for any input $\bm{s}=(x_0,\bm{y})$.
\end{corollary}

Finally, for end-to-end training, the gradients of  projection in Eqn.~\eqref{eqn:proj_ho} with respect to $\tilde{x}_t$ and $\text{cost}(x_{1:t-1})$ can be derived and the ML model can be trained following the steps in  Section~\ref{sec:training}. Hence, we omit them for brevity.

\section{Experimental Results}\label{sec:experiment} 

To empirically validate \ouralg, we consider the dynamic energy scheduling
application in the presence of uncertain renewables. 
 Specifically, renewable energy such as wind and solar energy is being massively incorporated into the power grid 
for sustainability. But, their availability is highly intermittent subject
to a variety of factors such as weather conditions and equipment efficiency. 
On the other hand, balancing the power demand and generation is crucial to ensure
grid stability --- a mismatch requires rapid offsetting using alternative and potentially more expensive energy sources. Thus, a challenging problem faced by grid operators
is \emph{how to
 dynamically schedule energy production to meet net demands based
on real-time renewable availability}. 
 A mismatch between the production $x_t$ and
net demand $y_t$ needs offsetting using expensive energy sources/storage and hence causes a \emph{hitting} cost $f(x_t,y_t)=\alpha\|x_t-y_t\|$, 
 and varying the production level over time incurs
 a \emph{memory} cost $d(x_t,x_{t-1})=\|x_t-x_{t-1}\|$ (due to generator ramp-up/-down costs). Thus, this is a typical online optimization problem with memory cost \cite{SOCO_Prediction_Error_RHIG_NaLi_Harvard_NIPS_2020_NEURIPS2020_a6e4f250,Shaolei_L2O_ExpertCalibrated_SOCO_SIGMETRICS_2022,SOCO_Revisiting_Nanjing_NIPS_2021_zhang2021revisiting}. 

\subsection{Dataset}
We consider intermittent renewable energy generated using trace data and empirical equations. Specifically, for wind power, the amount of energy generated at step $t$ is modeled based on \cite{wind_power_sarkar2012wind} as
$  P_{\mathrm{wind},t}=\frac{1}{2}\kappa_{\mathrm{wind}}\varrho A_{\mathrm{swept}} V_{\mathrm{wind},t}^3$.

The sympols are explained as follows:  $\kappa_{\mathrm{wind}}$ is the conversion efficiency (\%) of wind energy, $\varrho$ is the air density ($kg/m^3$), $A_{\mathrm{swept}}$ is the swept area of the turbine ($m^2$), and $V_{\mathrm{wind},t}$ is the wind speed ($kW/m^2$) at time step $t$. The amount of solar energy generated at step $t$ is given based on \cite{solar_power_wan2015photovoltaic} as 
 $   P_{\mathrm{solar},t}=\frac{1}{2}\kappa_{\mathrm{solar}} A_{\mathrm{array}} I_{\mathrm{rad},t}(1-0.05*(\mathrm{Temp}_t-25))$. 
The symbols are explained as follows: $\kappa_{\mathrm{solar}}$ is the conversion efficiency (\%) of the solar panel,  $A_{\mathrm{array}}$ is the array area ($m^2$), and $I_{\mathrm{rad},t}$ is the solar radiation ($kW/m^2$), and $\mathrm{Temp}_t$ is the temperature ($^{\circ}$C) at step $t$.
Thus, at time step $t$, the total energy generated by the renewables $P_{\mathrm{r},t}=P_{\mathrm{wind},t}+P_{\mathrm{solar},t}$.
 Suppose at time step $t$, the net energy demand is $y_t=\max(P_{\mathrm{s},t}-P_{\mathrm{r},t},0)$ , where $P_{\mathrm{s},t}$ is the demand before renewable integration.
The amount of energy generation is the agent's online action $x_t$.
We model the hitting cost as the scaled $l_2$-norm of the difference between the action $x_t$ and the context $y_t$, i.e. $f(x_t,y_t)=\alpha\|x_t-y_t\|$.
Additionally, we model the switching cost by the $l_2$-norm of the difference between two consecutive actions, i.e. $c(x_t,x_{t-1})=\|x_t-x_{t-1}\|$. 
The hitting cost parameter is set as $\alpha=0.2$.
The parameters for wind energy are set as $\kappa_{\mathrm{wind}}=30\%$, $\varrho=1.23 kg/m^3$, $A_{\mathrm{swept}}=500,000 m^2$. The parameters of solar energy are set as $\kappa_{\mathrm{wind}}=10\%$, $A_{\mathrm{array}}=10,000 m^2$. 
The other parameters, such as wind speed, solar radiation and temperature data, are all collected from the National Solar Radiation Database \cite{NSRDB_sengupta2018national}, which contains detailed hourly data for the year of 2015.

To generate datasets for training and testing, we use a sliding window to generate multiple sequences of 
hourly data, with each sequence length being 25 (i.e., 24 action steps plus 1 initial step). For each sequence
of 25 consecutive hourly data, we can calculate the contextual information for each step/hour. We define the energy generation of the first hour as the initial action $x_0$. 
The problem can be formulated as: 
$\min_{x_1,\cdots x_T} \sum_{t=1}^T \alpha \| x_t - y_t \| + \|x_t - x_{t-1} \|$. 
We use the CVXPY Library to find the optimal offline solution.

\subsection{Experimental Setup}
We use a RNN with 2 hidden layers, each with 8 neurons. To train this model, we use the data from the first two months (January--February) of 2015, which contains 1440 hourly weather data samples in total. Specifically, we generate 1416 data sequences using a sliding window. We train the RNN model for 140 epochs with batch size of 50. The model is implemented in PyTorch Library and the training process usually takes around 3 minutes on a 2020 MacBook Air with 8GB memory and a M1 chipset. 
 In \ouralg, we set the slackness parameter $B=0$ to follow the strict definition of competitive ratio.
By default, we train \ouralg with  $\lambda=1.4$. 

To evaluate the performance of different algorithms, we divide the remaining 10 months of 2015 into five 
segments, each with two months. 
There are three different cases:
when \ml empirically works better
than \greedy in terms of both average and worst-case performance;
\ml is better than \greedy on average but worse in the worst case;
and \ml is worse than \greedy both
on average and in the worst case.
The first case occurs for the testing segment of March--April, because the data in both training and testing datasets well consistent due to their similar weather patterns. Next,
we focus on the other two cases, which are more interesting and typical since data distributional shifts between training and testing datasets are very common in practice. This is also consistent with our main contribution --- robustifying ML-based optimizers.

While
we can also re-train/update the ML models
(in \ouralg, \ml, and \rob) based on
online collected data, the existing ML-based optimizers are typically pre-trained offline \cite{L2O_OnlineBipartiteMatching_Toronto_ArXiv_2021_DBLP:journals/corr/abs-2109-10380,L2O_OnlineResource_PriceCloud_ChuanWu_AAAI_2019_10.1609/aaai.v33i01.33017570}. Thus, we keep the ML model unchanged when testing its performance, in order to highlight the role of our expert robustification step in \ouralg --- regardless of the testing distributions, \ouralg offers a provable worst-case competitive ratio guarantee against the expert. 

\subsection{Baselines}

We compare \ouralg with the following baselines.
\textbf{Optimal offline (\opt):} \opt has all the context information to optimally solve Eqn.~\eqref{eqn:naive_online}; 
\textbf{Robust expert  (\greedy):} \greedy is the state-of-the-art expert
that chooses $x_t=\arg\min_{x\in\mathcal{X}}f(x,y_t)$ for  $t=1,\cdots,T$ with guaranteed competitive ratios \cite{SOCO_Revisiting_Nanjing_NIPS_2021_zhang2021revisiting}; 
\textbf{Simple greedy (\greedyswitch):} \greedyswitch
greedily minimizes the total hitting cost and memory cost at each step;
\textbf{Pure ML-based optimizer (\ml):} \ml 
uses the same recurrent neural network as \ouralg but does not use expert robustification for training or inference; 
\textbf{Dynamic switching  (\switch):} \switch
dynamically switches between \greedy and \ml based on  
a threshold hyperparameter \cite{SOCO_OnlineMetric_UntrustedPrediction_ICML_2020_DBLP:conf/icml/AntoniadisCE0S20}; 
\textbf{\ouralg-NoTraining (\rob):} \rob uses Algorithm~\ref{alg:EROP} for inference but
the ML model is trained as a standalone optimizer without end-to-end training.

Although \greedyswitch may empirically perform better than \greedy,
it does not have a provably-bounded competitive ratio whereas \greedy has
one (see \cite{SOCO_Revisiting_Nanjing_NIPS_2021_zhang2021revisiting}
for the single-step memory case
and our Theorem~\ref{thm:greedy_multi} for the multi-step memory case). Thus,
we use \greedy as our expert in \ouralg.

\subsection{Results}

\begin{figure}[t!]
  \subfigure[September-December testing]{\includegraphics[trim=0 0cm 0 0, clip, width=0.24\textwidth]{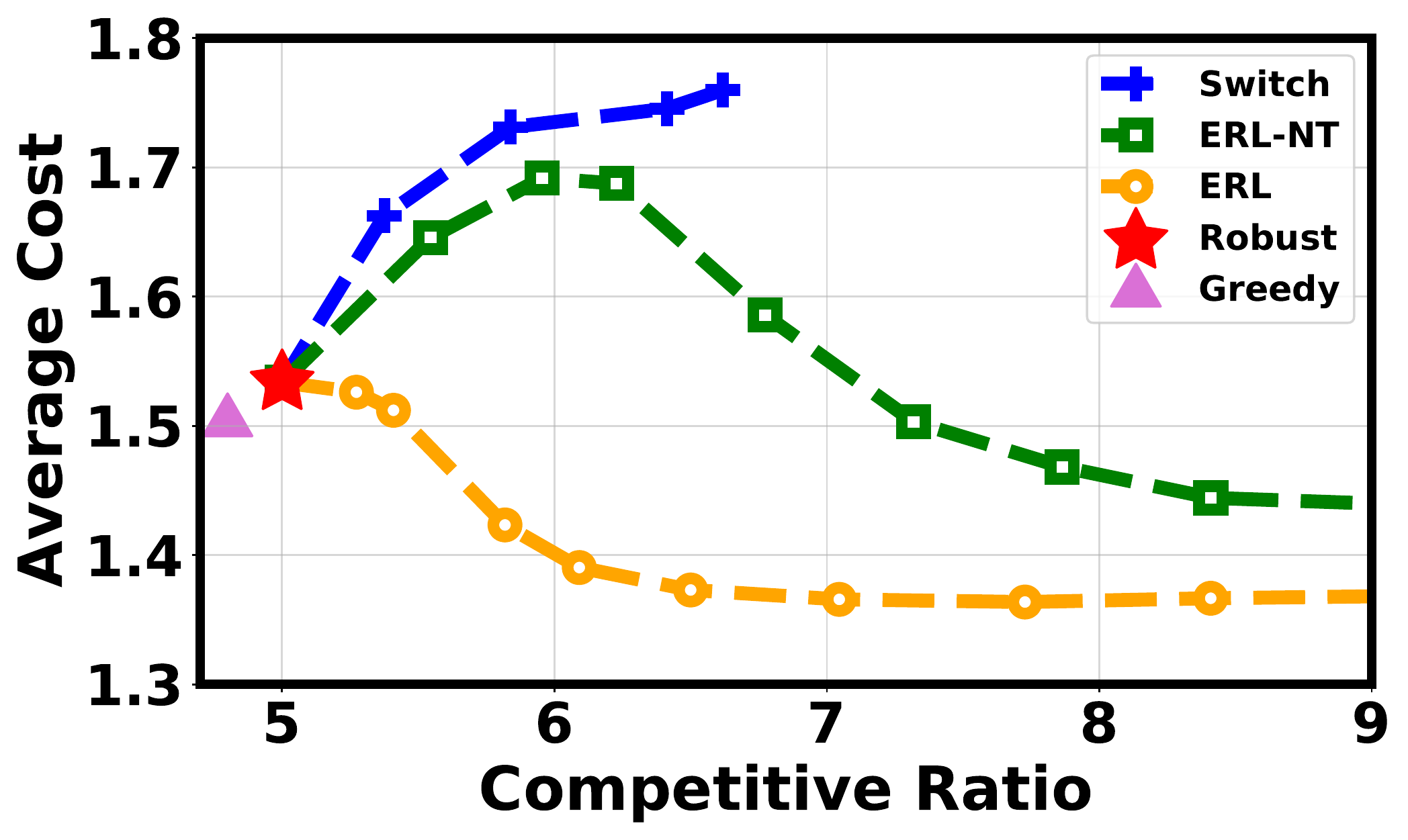}\label{fig:avg_cr_main}}
  \subfigure[May-August testing]{\includegraphics[trim=0 0cm 0 0, clip, width=0.24\textwidth]{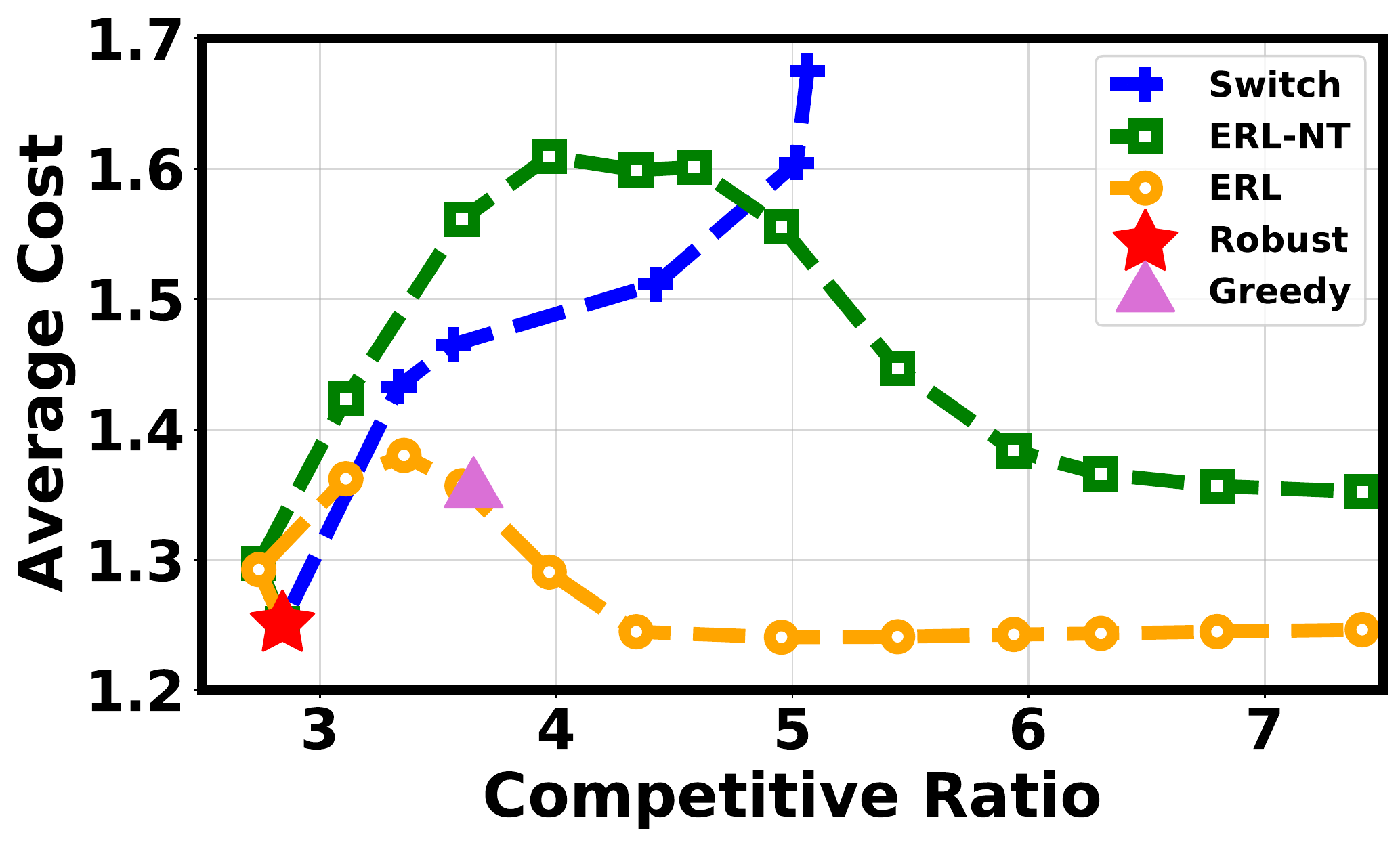}\label{fig:tradeoff_large_ood}}
  \vspace{-0.3cm}
    \caption{Normalized average cost vs. empirical competitive ratio. 
    \ml is off the charts: (left): average cost 1.45 and competitive ratio 50+;  (right): average cost 1.367 and competitive ratio 11.167.}
    \label{fig:results}
\end{figure}

\begin{figure*}[t!]
    \centering 
    \subfigure[]{\includegraphics[width=0.24\textwidth]{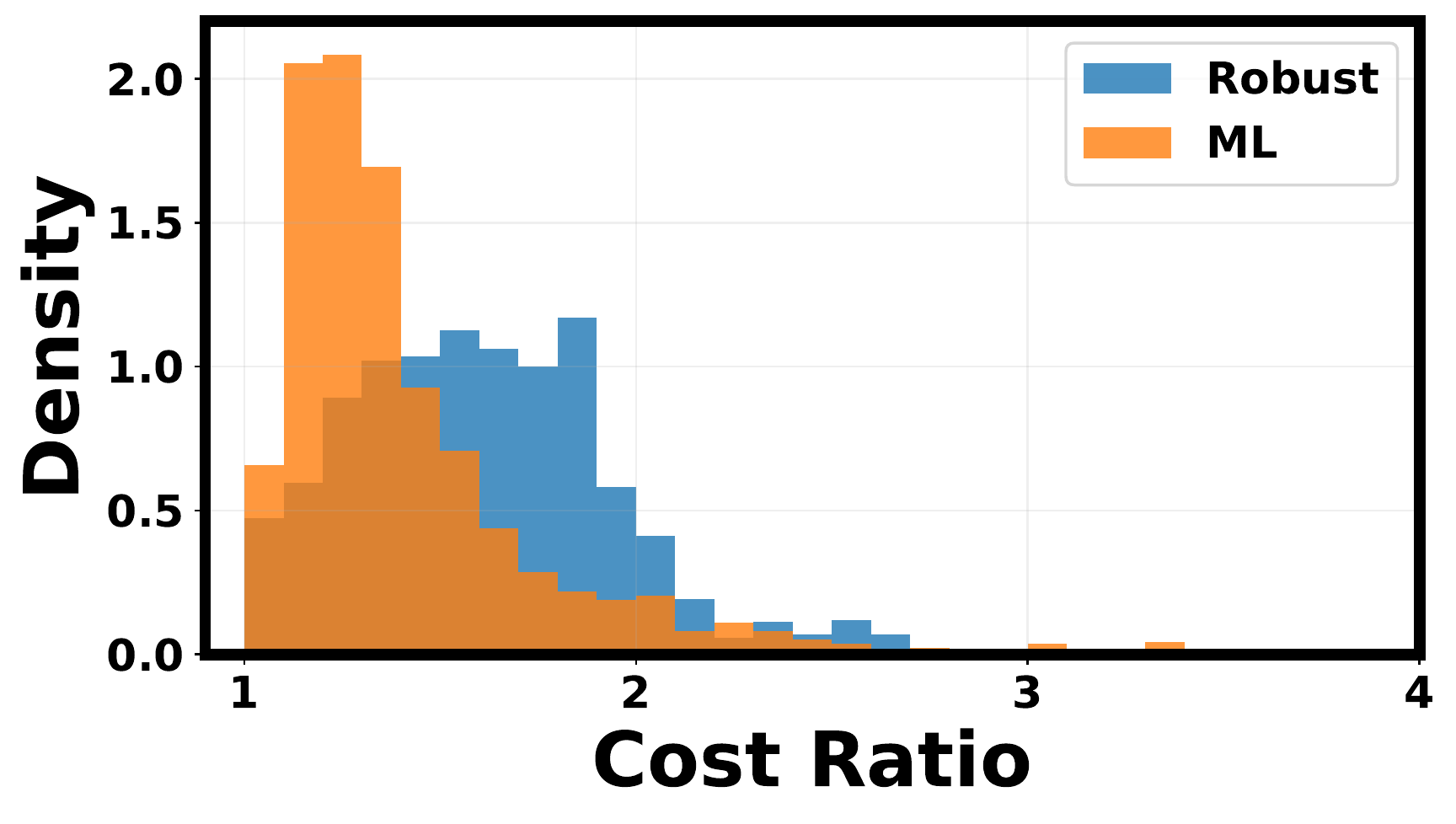}\label{fig:sub_image0}}
\subfigure[]{\includegraphics[width=0.24\textwidth]{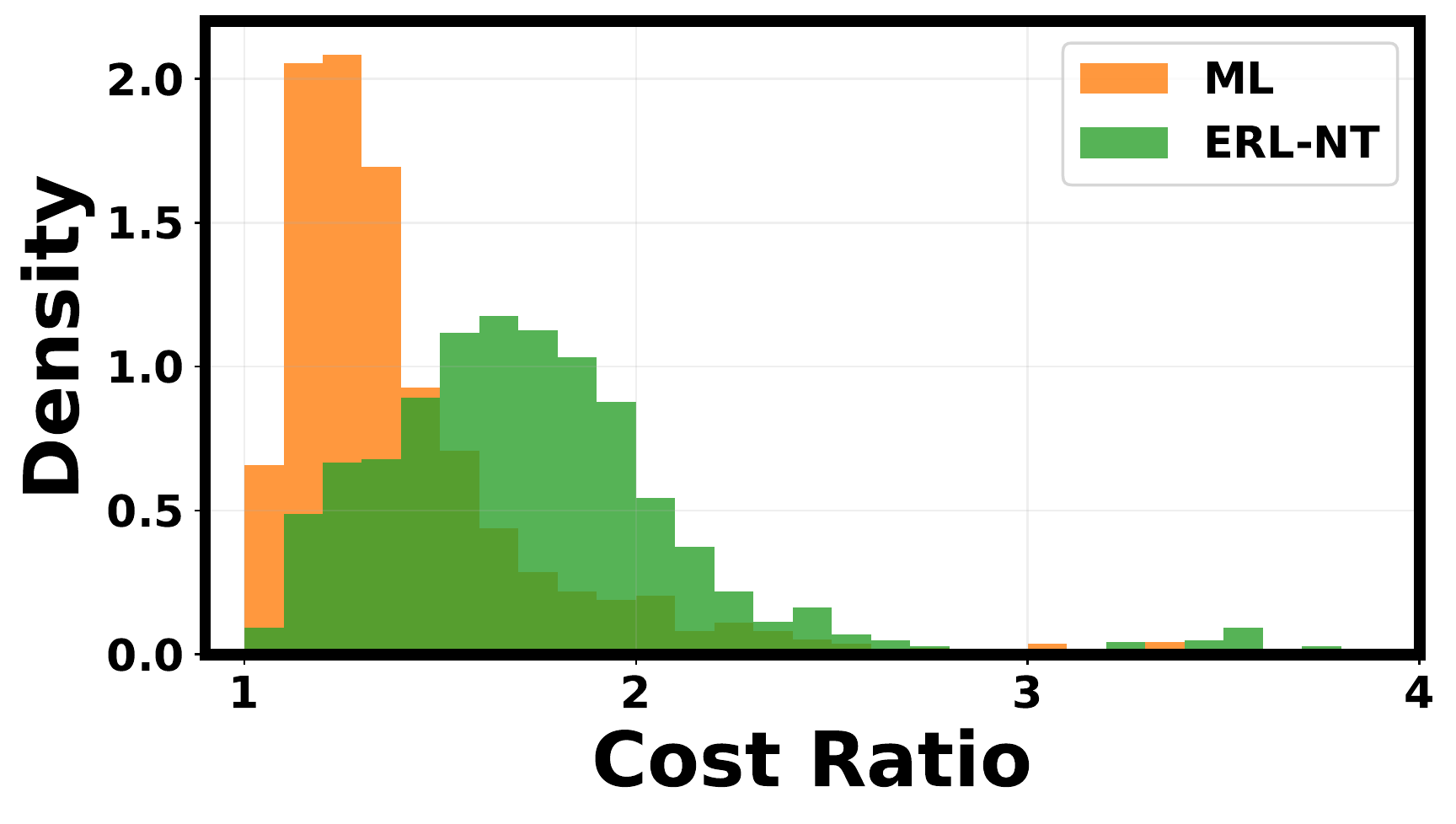}\label{fig:sub_image1}}
\subfigure[]{\includegraphics[width=0.24\textwidth]{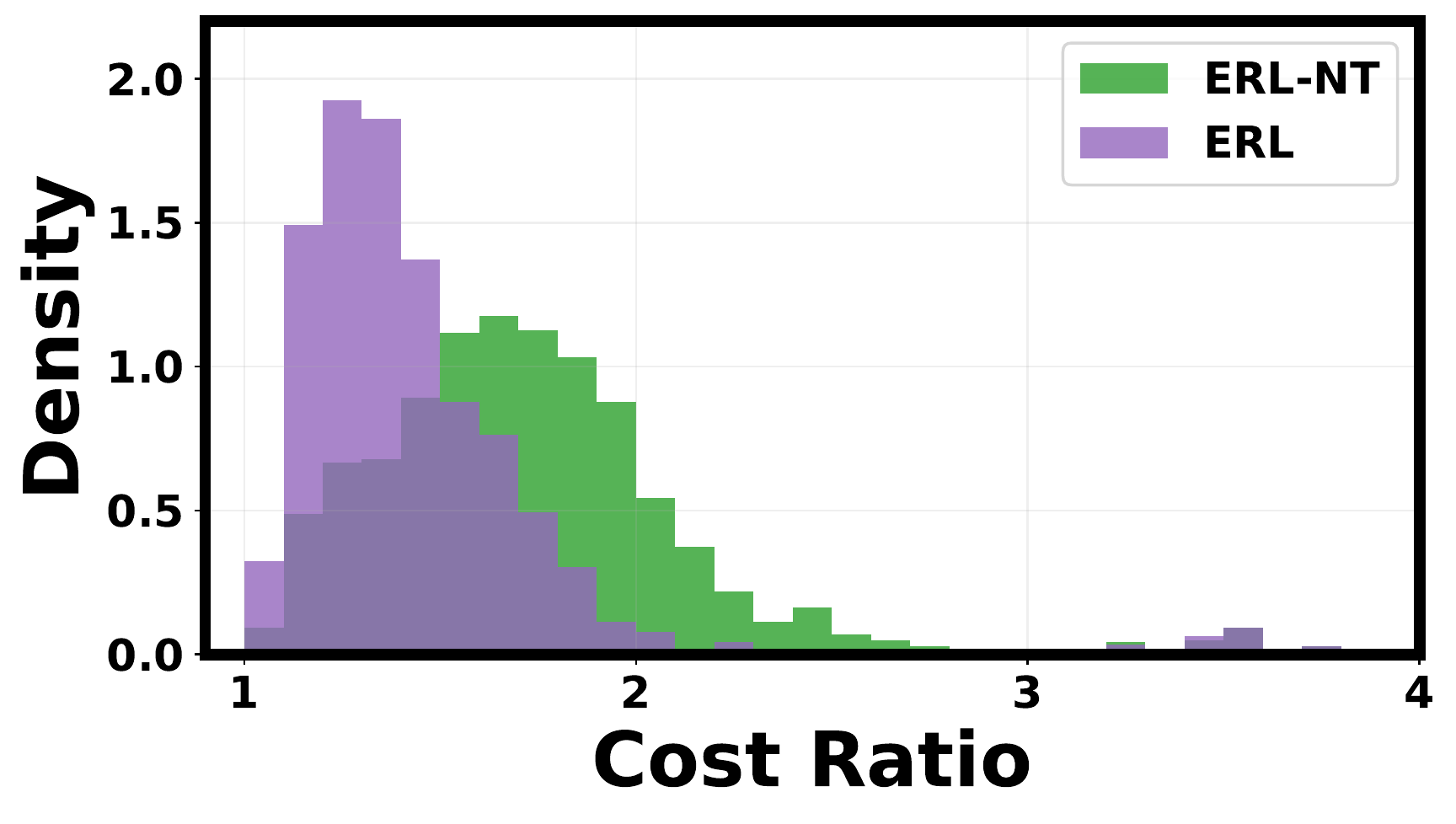}\label{fig:sub_image2}} 
\subfigure[]{\includegraphics[width=0.24\textwidth]{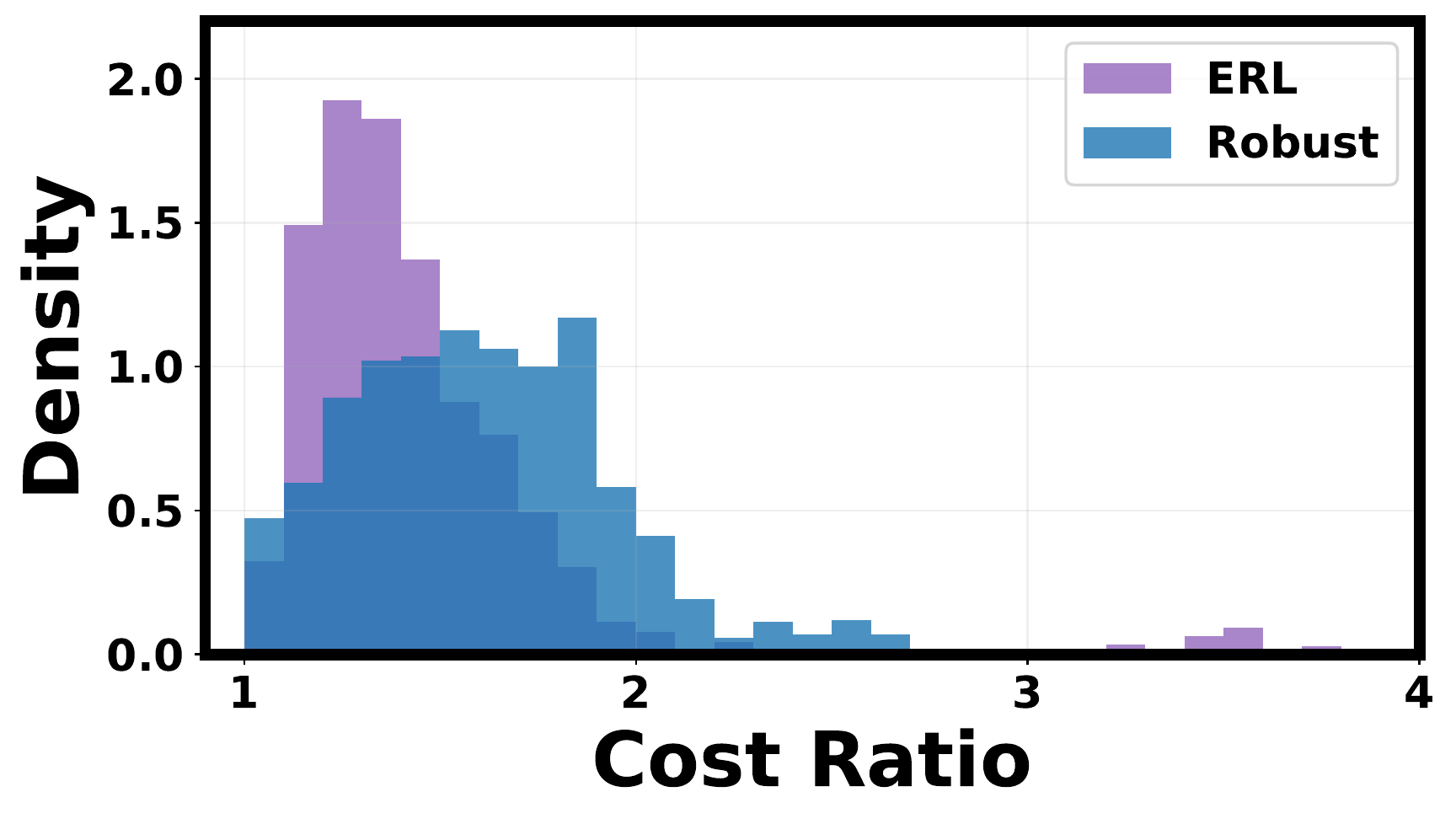}\label{fig:sub_image4}} 
\vspace{-0.3cm}
    \caption{Cost ratio probability distribution for September-December testing. 
    The density values are obtained by
    dividing the number of testing instances
    within each bin by the total number
    of instances and the bin width.
    We use $\lambda=1.4$ for testing \rob and \ouralg. The horizontal axis is limit to $4$ for better visualization, and  cost ratios larger than 4 are off the charts.}
    \label{fig:PMF_dist}
\end{figure*}

\textbf{September-December testing.}
We obtain the empirical results
of average cost vs. competitive ratio in Fig.~\ref{fig:results}.
All the average costs are normalized with respect to the average cost of \opt. 
\ml achieves a lower average cost than \greedy, but its empirical competitive ratio is way larger due to the common drawback of ML-based optimizers --- lack of performance robustness. Specifically,
the training and testing distributions are rarely identical in practice, which can lead to an extremely bad competitive ratio for \ml.
While \greedyswitch empirically performs better than \greedy in this setting,
it does not have any competitive ratio guarantees.
We see
that \switch performs badly compared to \greedy, because it imposes a hard switch based on a pre-set threshold regardless of the actual performance of \greedy or \ml. Compared to \ml,
\rob can have a much lower competitive ratio due to expert robustification, but the average cost also increases dramatically
and can be even higher than \greedy (because the ML model
training in
\rob is not aware of the robustification step). On the other hand, \ouralg
achieves a guaranteed competitive ratio
and a much lower average cost than \rob. This highlights the benefits of training the ML model in \ouralg by explicitly considering the downstream expert robustification process. Interestingly, we also observe that by properly setting the  hyperparameter $\lambda$ (around $1.4\sim1.8$ in our case), \ouralg can have an even lower average cost than  \ml. This is because for those \emph{hard} problem instances that \ml cannot solve well,
\ouralg has \greedy as its guidance to provide reasonably good solutions.

\textbf{Cost ratio distribution.}  To provide further insights, we also show in Fig.~\ref{fig:PMF_dist} the detailed comparison between different algorithm pairs
in terms of the cost ratio distribution density. 
By looking at \greedy vs. \ml in Fig.~\ref{fig:sub_image0}, we can see that \ml has low cost ratios in more cases than \greedy, although it has a long tail (not shown in the figure due to the axis limit). This explains that \ml can have good average performance than the expert algorithm \greedy, when the training-testing distributions are not very different. Nonetheless, \ml still suffers from the lack of robustness, while \greedy does not.
Comparing \ml with \rob in Fig.~\ref{fig:sub_image1}, we can see that expert robustification can shift the cost ratios rightwards (i.e., increasing the average cost), but \rob has guaranteed robustness. 
Next, we observe from Fig.~\ref{fig:sub_image2}
that the cost ratios of \ouralg are shifted leftwards compared to \rob, demonstrating the importance of training \ouralg with explicit consideration of the expert robustification process. 
Fig.~\ref{fig:sub_image4} shows that \ouralg has
many smaller cost ratios than \greedy.
Again, this shows the importance of considering
expert robustficniation during the training process.

\begin{table*}[t!]
\centering
\begin{tabular}{l|ll|ll|ll|ll|ll|ll} 
\toprule
\multicolumn{1}{l}{}  & \multicolumn{2}{c}{\rob}& \multicolumn{2}{c}{\ouralg ($\lambda$ = 1.4)}                   & \multicolumn{2}{c}{\ouralg ($\lambda$ = 1.2)}                   & \multicolumn{2}{c}{\switch}  & \multicolumn{2}{c}{\greedy}  & \multicolumn{2}{c}{\greedyswitch}\\ 
\hline
\multicolumn{1}{c|}{$\lambda$ for Testing} & \multicolumn{1}{c}{Avg} & \multicolumn{1}{c|}{CR} & \multicolumn{1}{c}{Avg} & \multicolumn{1}{c|}{CR} & \multicolumn{1}{c}{Avg} & \multicolumn{1}{c|}{CR} & \multicolumn{1}{c}{Avg}   & \multicolumn{1}{c|}{CR}   & \multicolumn{1}{c}{Avg}   & \multicolumn{1}{c|}{CR}  
& \multicolumn{1}{c}{Avg}   & \multicolumn{1}{c}{CR}    \\ 
\hline
$\lambda$ = 1.4 & 1.6977 & 6.0912 & \textbf{1.3903}    & {6.0910}   & 1.4343    & 6.0910   & 1.7454 & 6.4130 & 1.5336 & 5.000 & 1.5030 & 4.800\\
$\lambda$ = 1.2          & 1.6457    & 5.5457   & 1.4832    & 5.5456   & \textbf{1.4587}    & {5.5456}   & 1.7454 & 6.4130    & 1.5336 & 5.000 
& 1.5030 & 4.800\\ 
\bottomrule
\end{tabular}
\vspace{-0.1cm} 
\caption{September-December testing. ``Avg'' and ``CR'' represent the empirical average cost and competitive ratio (normalized w.r.t. \opt), respectively. 
Bold texts mean the best AVG performance.
\ouralg ($\lambda=x$) means we train $\ouralg$ with $\lambda=x$.}
\label{tab:different_lambda}
\end{table*}
\textbf{Impacts of $\lambda.$} While
both \ouralg and \rob can guarantee robustness due to the expert robustification step during inference,
the ML model in \ouralg is trained with explicit consideration of the expert whereas \rob simply trains the ML model as a standalone optimizer.
Thus, \ouralg can further improve the average performance compared to \rob.
 To further highlight the necessity
of being aware of the expert robustification step in the training of \ouralg, 
we show the results for different algorithms in  Table~\ref{tab:different_lambda}.
By training \ouralg 
using the same $\lambda$ as testing it,
we can obtain both the best average cost and the best competitive ratio empirically. In particular, the difference in terms of the average performance is more prominent when $\lambda=1.4$ than when $\lambda=1.2$.
This can be explained by noting that with a larger $\lambda\geq1$, the expert plays a less significant role by placing less emphasis on robustness and providing the ML model with more freedom. Then, when $\lambda=1.4$,
the average performance is better than when $\lambda=1.2$, although its guaranteed competitive ratio is  higher (which is also empirically verified in Table~\ref{tab:different_lambda}).
For reference, we also show the performance of other algorithms
that are not affected by $\lambda\geq1$.

\textbf{May-August testing.} Next, we turn to a more challenging case
in which \ml is outperformed by  \greedy
both on average and in the worst case
(May--August, due to the different weather patterns and hence large training-testing distributional shifts). This is not uncommon in practice, since ML models can have arbitrarily bad performances due to the lack of robustness.
We show the results
in Fig.~\ref{fig:tradeoff_large_ood}.
All the average values are normalized with respect to the  average cost of \opt. 

Again, \ml is  off the charts, with its competitive ratio as 11.167 and average cost as 1.367 (both normalized with respect to \opt). 
Like in the previous case, \switch is 
not as good as \greedy, since it utilizes a \emph{hard} switching between \greedy and \ml whenever a pre-defined threshold is reached without looking at the actual performance of \greedy or \ml. 
Due to the lack of robustness guarantees, \greedyswitch is also worse than \greedy in this setting.
By varying $\lambda\geq1$, we see
that \rob can have very large average costs (even larger than \ml), although its competitive ratio is still guaranteed to be $\lambda$-competitive against the expert \greedy. 
On the other hand, \ouralg, which is trained with $\lambda=1.4$ and tested
with different $\lambda\geq1$ has a much lower average cost than \rob, while also being able to guarantee competitive ratios.  This shows the importance of being aware of expert robustification during
the training stage.
Moreover, \ouralg has a lower average
cost than \ml: even in the presence
of large training-testing distributional discrepancies, the expert can help correct many of the bad pre-robustification actions, thus significantly improving the average performance of \ouralg over \ml. Interestingly, the average performance of \ouralg is not monotonic in
the parameter of $\lambda\geq1$ used for testing. This is partly because $\lambda$ is different for training and testing, and partly because the large training-testing distributional discrepancies result
in irregular average performance
for the ML model used by \ouralg. By $\lambda=1$, we essentially have no trust on the ML model in \ouralg, and hence \ouralg will follow the expert \greedy at each step.

\textbf{Summary.}
Our experiments highlight the key point that \ouralg guarantees worst-case robustness in terms of the competitive ratio by utilizing expert robustification, while exploiting the power of ML to improve the average performance. Naturally, when training-testing distributions are reasonably similar, we expect the average performance of \ouralg (and other ML-based optimizers like \ml) to be better than that of \greedy. But, even when the pure \ml performs arbitrarily badly, \ouralg can still offer a good average cost performance due to the introduction of expert robustification. Last but not least, with explicit awareness of the expert robustification process, 
\ouralg has a much better average performance than otherwise
(i.e., \rob).

\section{Conclusion}

In this paper, we propose \ouralg, a novel expert-robustified learning  approach to solve
online optimization with memory costs.
For guaranteed robustness, \ouralg
introduces a projection operator that
robustifies ML actions by utilizing an expert online algorithm;
for good average performance, \ouralg
trains the ML optimizer based on a recurrent architecture by explicitly considering
downstream expert robustification process. 
We prove that, for any $\lambda\geq1$,
\ouralg can achieve $\lambda$-competitive against the expert algorithm for any problem inputs.
We also extend our analysis to a novel setting of multi-step memory costs.
Finally, we run experiments for an energy scheduling application to validate \ouralg, showing that \ouralg
 can offer the best tradeoff in terms of the average and worst performance.

\section*{Acknowledgement} 
 This work was supported in part by the NSF under grant CNS-1910208.
\appendix
In the more general case, the memory cost may span multiple steps (e.g. acceleration smoothness),
which has not been well studied. 
 We first show that \greedy
is still an competitive expert, by  
providing its competitive ratio in the multi-step memory setup in Appendix~\ref{sec:greedy_porrf}. Then, in Appendix~\ref{sec:proof_E2}, we prove that \ouralg is still $\lambda$-competitive against any expert,
and this automatically proves Theorem~\ref{theorem:cr_online}
for the single-stem memory case.

\subsection{Proof of Theorem \ref{thm:greedy_multi}}\label{sec:greedy_porrf}

When $t\geq q$, \greedy satisfies the following condition:

\begin{equation}\label{eq1}\nonumber
\begin{split}
 & f(x_t^\pi, y_t) + ||x_t^\pi - \sum_{i=1}^q C_i x_{t-i}^\pi || \\
 \leq &   f(x_t^\pi, y_t) + ||x_t^* - \sum_{i=1}^q C_i x_{t-i}^* || + ||x_t^\pi - x_t^*||\\
 &+ \sum_{i=1}^q ||C_i||\cdot ||x_{t-i}^\pi - x_{t-i}^*|| \\
 \leq &   f(x_t^\pi, y_t) + ||x_t^* - \sum_{i=1}^q C_i x_{t-i}^* || + \frac{1} {\alpha} \big( f(x_t^*, y_t ) - f(x_t^\pi, y_t ) \big)\\
 & + \frac{1} {\alpha} \sum_{i=1}^q ||C_i|| \big( f(x_{t-i}^*, y_{t-i} ) - f(x_{t-i}^\pi, y_{t-i} ) \big). 
\end{split}
\end{equation}

The first and second inequalities come from the triangle inequality of $l_p$ norm, and the third inequality comes from the $
\alpha$-polyhedral assumption of the hitting cost function. For $t < q$, since $x_t^\pi = x_t^*=x_t, \forall t \in [-q+1, 0]$, the above inequality also holds. 
We sum up all the single-step costs:

\begin{equation} \label{eqn:greedyproof1}
\nonumber
\begin{split}
 & \sum_{t = 1}^T f(x_t^\pi, y_t) + \left\|x_t^\pi - \sum_{i=1}^q C_i x_{t-i}^\pi \right\| \\
 \leq &   \sum_{t = 1}^T f(x_t^\pi, y_t) + \sum_{t = 1}^T \left\|x_t^* - \sum_{i=1}^q C_i x_{t-i}^* \right\|\\
 &+ \frac{1} {\alpha} \sum_{t = 1}^T \big( f(x_t^*, y_t ) - f(x_t^\pi, y_t ) \big)\\
 & + \frac{1} {\alpha}  \sum_{i=1}^q \|C_i\| \sum_{t = 1}^T \big( f(x_{t-i}^*, y_{t-i} ) - f(x_{t-i}^\pi, y_{t-i} ) \big) \\
 \leq &   \sum_{t = 1}^T f(x_t^\pi, y_t) + \sum_{t = 1}^T \left\|x_t^* - \sum_{i=1}^q C_i x_{t-i}^*\right\|\\
 &+ \frac{1} {\alpha} \sum_{t = 1}^T \big( f(x_t^*, y_t ) - f(x_t^\pi, y_t ) \big)\\
 & + \frac{1} {\alpha}  \sum_{i=1}^q \|C_i\| \sum_{t = 1}^T \big( f(x_{t}^*, y_{t} ) - f(x_{t}^\pi, y_{t} ) \big)\\
 = & \sum_{t = 1}^T f(x_t^\pi, y_t) + \sum_{t = 1}^T \left\|x_t^* - \sum_{i=1}^q C_i x_{t-i}^* \right\| \\
 &+ \frac{1} {\alpha}(1+ \sum_{i=1}^q \|C_i\|)\sum_{t = 1}^T \big( f(x_t^*, y_t ) - f(x_t^\pi, y_t ) \big),
\end{split}
\end{equation}

where the second inequality holds because $x_t^\pi = x_t^*=x_t, \forall t \in [-q+1, 0]$ and $f(x_t^*, y_t ) - f(x_t^\pi, y_t )\geq 0$.  
Thus, we have
\begin{equation} \label{eqn}
\begin{split}
 & \sum_{t = 1}^T f(x_t^\pi, y_t) + \left\|x_t^\pi - \sum_{i=1}^q C_i x_{t-i}^\pi \right\| \\
 \leq & (1 - \frac{1+ \beta}{\alpha}) \sum_{t = 1}^T f(x_t^\pi, y_t) + \sum_{t = 1}^T \left\|x_t^* - \sum_{i=1}^q C_i x_{t-i}^* \right\| \\
 +& \frac{1+ \beta} {\alpha}\sum_{t = 1}^T f(x_t^*, y_t )
\end{split}
\end{equation}

If $\alpha \leq 1+\beta$, then $1 - \frac{1+ \beta}{\alpha} \leq 0$, the inequality~\eqref{eqn} becomes
\begin{equation}\nonumber\begin{split}
&\sum_{t = 1}^T f(x_t^\pi, y_t) + \left\|x_t^\pi - \sum_{i=1}^q C_i x_{t-i}^\pi \right\| \\
\leq&  \frac{1+ \beta} {\alpha}\sum_{t = 1}^T f(x_t^*, y_t ) + \sum_{t = 1}^T \left\|x_t^* - \sum_{i=1}^q C_i x_{t-i}^\pi \right\|.
\end{split}
\end{equation}
If $\alpha > 1+\beta$, since $x_t^\pi = v_t$ minimizes $f(\cdot, y_t)$, then $f(x_t^\pi, y_t) \leq f(x_t^*, y_t)$ and,
based on 
\eqref{eqn}, \greedy is optimal. This completes the proof.

\subsection{Proof of Theorem~\ref{theorem:cr_online}
and Corollary~\ref{corollary:cr_opt_multi}}\label{sec:proof_E2} 
We denote the accumulated cost of the first $t_1$ steps as 
$\mathrm{cost}({x}_{1:t_1}) = \sum_{t=1}^{t_1} \big( f(x_t, y_t) + \tilde{d}(x_t , \sum_{i=1}^q C_i x_{t-i}) \big)$.
When $t=1$, $x^{\pi}_1$ is clearly a  feasible solution to \eqref{eqn:proj_ho}. 
Then, suppose that for $t \geq 1$, $x_{1:t-1}$ satisfies the constraint, i.e.
$\mathrm{cost}(x_{1:t-1}) + G(x_{t-1}, x_{t-q-1:t-2}, x_{t-q-1:t-1}^\pi) -  \big(\lambda\mathrm{cost}(x_{1:t-1}^\pi) + B \big) \leq  0.$
We need to prove that $x_t^{\pi}$ is a feasible solution of the projection ~\eqref{eqn:proj_ho}. By the constraint in
the projection, we have
\begin{equation}\label{eqn:hoproof_2}\nonumber
    \begin{split} 
     & \big( \mathrm{cost}(x_{1:t-1}) + f(x_t^\pi, y_t) + \tilde{d}(x_t^\pi,  \sum_{i=1}^q C_i x_{t-i}) \big) \\
     &+ G(x_t^\pi, x_{t-q:t-1}, x_{t-q:t}^\pi) -  \big(\lambda\mathrm{cost}(x_{1:t}^\pi) + B \big) \\
     = & \mathrm{cost}(x_{1:t-1}) - \big(\lambda\mathrm{cost}(x_{1:t-1}^\pi) + B \big) \\
     &+ G(x_t^\pi, x_{t-q:t-1}, x_{t-q:t}^\pi)\\
     &+ \tilde{d}(x_t^\pi,  \sum_{i=1}^q C_i x_{t-i}) - \tilde{d}(x_t^\pi,  \sum_{i=1}^q C_i x_{t-i}^\pi).
    \end{split}
\end{equation}
By the triangular inequality, we have
$\tilde{d}(x_t^\pi,  \sum_{i=1}^q C_i x_{t-i}) - \tilde{d}(x_t^\pi,  \sum_{i=1}^q C_i x_{t-i}^\pi) \leq  \tilde{d}(\sum_{i=1}^q C_i x_{t-i}, \sum_{i=1}^q C_i x_{t-i}^\pi)$ and 
\begin{equation}
    \begin{split} 
     & \big( \mathrm{cost}(x_{1:t-1}) + f(x_t^\pi, y_t) + \tilde{d}(x_t^\pi,  \sum_{i=1}^q C_i x_{t-i}) \big)\\
     &+ G(x_t^\pi, x_{t-q:t-1}, x_{t-q:t}^\pi) -  \big(\lambda\mathrm{cost}(x_{1:t}^\pi) + B \big) \\
     \leq & \mathrm{cost}(x_{1:t-1}) - \big(\lambda\mathrm{cost}(x_{1:t-1}^\pi) + B \big) \\
     &+ G(x_t^\pi, x_{t-q:t-1}, x_{t-q:t}^\pi)\\
     &+\tilde{d}(\sum_{i=1}^q C_i x_{t-i}, \sum_{i=1}^q C_i x_{t-i}^\pi)\\
     = & \mathrm{cost}(x_{1:t-1}) - \big(\lambda\mathrm{cost}(x_{1:t-1}^\pi) + B \big)\\& +\sum_{k = 1}^{q}{\tilde{d}(\sum_{i=1}^{q-k} C_{k + i}x_{t-i}, \sum_{i=1}^{q-k} C_{k + i}x_{t-i}^\pi)}\\
    & + \tilde{d}(\sum_{i=1}^q C_i x_{t-i}, \sum_{i=1}^q C_i x_{t-i}^\pi)\\
     = &\mathrm{cost}(x_{1:t-1}) + G(x_{t-1}, x_{t-q-1:t-2}, x_{t-q-1:t-1}^\pi)\\
     &-  \big(\lambda\mathrm{cost}(x_{1:t-1}^\pi) +B \big)
     \leq  0,
    \end{split}
\end{equation}
where the first inequality is because of the triangular inequality of $l_p$ norm, the first equality is from 
\begin{equation}
    \begin{split} 
     &G(x_t^\pi, x_{t-q:t-1}, x_{t-q:t}^\pi)\\
     =& \sum_{k = 1}^{q}{\tilde{d}(\sum_{i=1}^{q-k} C_{k + i}x_{t-i}, \sum_{i=1}^{q-k} C_{k + i}x_{t-i}^\pi)},
    \end{split}
\end{equation} and the second equality is from
\begin{equation}
    \begin{split} 
    & \sum_{k = 1}^{q}{\tilde{d}(\sum_{i=1}^{q-k} C_{k + i}x_{t-i}, \sum_{i=1}^{q-k} C_{k + i}x_{t-i}^\pi)} \\
    &+ \tilde{d}(\sum_{i=1}^q C_i x_{t-i}, \sum_{i=1}^q C_i x_{t-i}^\pi)\\
    =& \sum_{k = 0}^{q}{\tilde{d}(\sum_{i=1}^{q-k} C_{k + i}x_{t-i}, \sum_{i=1}^{q-k} C_{k + i}x_{t-i}^\pi)}\\
    =&  G(x_{t-1}, x_{t-q-1:t-2}, x_{t-q-1:t-1}^\pi)
    \end{split} 
\end{equation}
Thus, Theorem~\ref{theorem:cr_online} is proved by
setting $q=1$. 
By Theorem~\ref{thm:greedy_multi}, we also prove Corollary \ref{corollary:cr_opt_multi}.

\clearpage
\bibliographystyle{plain}
\bibliography{main_infocom.bbl}

\end{document}